%% file: _main.tex
\documentclass[runningheads]{llncs}

\usepackage{eccvabbrv}

\usepackage{eccv}
\input{aux/preamble.tex}

\usepackage[pagebackref,breaklinks,colorlinks, citecolor=eccvblue]{hyperref}
\usepackage{orcidlink}
\usepackage[capitalize]{cleveref}
\crefname{section}{Sect.}{Sects.}
\Crefname{section}{Section}{Sections}
\Crefname{table}{Table}{Tables}
\crefname{table}{Tab.}{Tabs.}

\begin{document}
\title{Cross-Attentive Multiview Fusion of Vision-Language Embeddings}
\titlerunning{CAMFusion}
\author{\authorBlock}
\authorrunning{T.~Berriel Martins et al.}

\maketitle
\input{sections/00_abstract.tex}
\input{sections/01_introduction.tex}
\input{sections/02_realted_work.tex}
\input{sections/03_method.tex}
\input{sections/04_experiments.tex}
\input{sections/10_conclusion.tex}

{\small
\bibliographystyle{splncs04}
\bibliography{main.bib}
}
\clearpage
\input{sections/11_appendix} 

\end{document}

%% file: aux/preamble.tex
\usepackage[dvipsnames,table]{xcolor} 
\usepackage{booktabs}
\usepackage{graphicx}
\usepackage{diagbox}
\usepackage{multirow}
\usepackage{amsmath}
\usepackage{amssymb} 
\usepackage{arydshln} 
\usepackage{pifont}
\usepackage{color, colortbl}
\usepackage{makecell}
\usepackage{comment}
\usepackage{url}

\usepackage{times}
\usepackage{microtype}
\usepackage{epsfig}
\usepackage{caption}
\usepackage{float}
\usepackage{placeins}
\usepackage{stfloats}
\usepackage{enumitem}
\usepackage{xstring}
\usepackage{xspace}
\usepackage{subcaption}
\usepackage{tabularx}
\usepackage[hang,flushmargin]{footmisc}

\usepackage[accsupp]{axessibility}  
\usepackage{pgfplots,tikz}
\pgfplotsset{compat=1.18}
\newcommand{\tablesize}{
\footnotesize
}
\colorlet{colorFst}{Green!25}       
\colorlet{colorSnd}{SpringGreen!45} 
\colorlet{colorTrd}{Yellow!30}      
\colorlet{colorLow}{darkgray!30}    
\newcommand{\st}{\cellcolor{colorFst}\bf}   
\newcommand{\nd}{\cellcolor{colorSnd}}      
\newcommand{\rd}{\cellcolor{colorTrd}}      

\definecolor{shadecolor}{rgb}{0.8,0.8,0.8}

\DeclareTextFontCommand{\bemph}{\bfseries\em}

\def\authorBlock{
    Tomás Berriel Martins\inst{1}
    \and
    Martin R. Oswald\inst{2}
    \and
    Javier Civera\inst{1}
}

\institute{University of Zaragoza, Spain \and
University of Amsterdam, Netherlands
\\
}

\def\ours{CAMFusion}

%% file: sections/00_abstract.tex
\begin{abstract}
Vision-language models have been key to the development of open-vocabulary 2D semantic segmentation. Lifting these models from 2D images to 3D scenes, however, remains a challenging problem. Existing approaches typically back-project and average 2D descriptors across views, or heuristically select a single representative one, often resulting in suboptimal 3D representations.
In this work, we introduce a novel multiview transformer architecture that cross-attends across vision-language descriptors from multiple viewpoints and fuses them into a unified per-3D-instance embedding. 
As a second contribution, we leverage multiview consistency as a self-supervision signal for this fusion, which significantly improves performance when added to a standard supervised target-class loss.
Our \underline{C}ross-\underline{A}ttentive \underline{M}ultiview \underline{Fusion}, which we denote with its acronym \textbf{\ours}, not only consistently outperforms naive averaging or single-view descriptor selection, but also achieves state-of-the-art results on 3D semantic and instance classification benchmarks, including zero-shot evaluations on out-of-domain datasets.
\end{abstract}

%% file: sections/01_introduction.tex
\section{Introduction}
\input{figures/11_teaser}
Semantic understanding is a key enabler for a wide range of applications, from autonomous robotic interaction~\cite{ewen2024feelit} and navigation~\cite{lee2025terrain} to long-term scene monitoring~\cite{schmid2022panoptic} and augmented reality~\cite{stanescu2023state}. Traditional methods, however, are often constrained by strong inductive biases and closed-set semantic formulations, which require manual tuning, costly retraining or additional data collection to accommodate new, previously unseen concepts in real-world scenarios. 

Recent advances in vision–language models (VLMs) have demonstrated remarkable capabilities for open-vocabulary detection and segmentation~\cite{zhu2024survey}. Pre-trained on large-scale datasets, VLMs align visual and textual representations in a shared high-dimensional space, enabling recognition and reasoning about novel objects without explicit supervision. This capability is particularly valuable in unstructured, in-the-wild settings, where the set of possible objects is effectively unbounded.

Although significant progress has been made in extracting rich open-vocabulary descriptors from single 2D views~\cite{radford2021learning,zhou2022extract,cherti2023reproducible,Zhai_2023_ICCV,sun2024alpha,zhang2024long}, a critical gap remains in effectively extending them to multiview 3D reconstructions. The absence of large-scale 3D-language datasets hinders the direct training of 3D-language models. As a result, current approaches predominantly rely on simple aggregation strategies to fuse descriptors from multiple views, such as average pooling~\cite{peng2023openscene,nguyen2024open3dis,Nguyen_2025_CVPR,Werby-RSS-24, koch2024open3dsg} or heuristic selection of a single descriptor to represent all views~\cite{werby2024hierarchical, ovo}. This, however, oversimplifies the vision-language embedding, and often fail to take advantage of the unique information present in different views—such as a view that reveals a handle versus a view that reveals a spout. Multiview vision-language encodings should not be treated as redundant, noisy measurements to be averaged out, but rather as complementary perspectives of a richer semantic whole.

In this work, we address this limitation by re-formulating multiview descriptor fusion as a \textit{semantic super-resolution} task instead of as a simplistic aggregation of features. Drawing inspiration from advances in novel view synthesis~\cite{Sajjadi_2022_CVPR} and multiview models~\cite{cabon2025must3r}, we adapt a multiview transformer architecture to alternate between each viewpoint, self-attending to itself and cross-attending to other viewpoints, enabling the model to actively aggregate complementary semantic details into a unified, high-fidelity descriptor. We will refer to this \underline{C}ross-\underline{A}ttentive \underline{M}ultiview \underline{Fusion} architecture by the acronym \textbf{\ours}.

In addition, to ensure robust generalization and prevent the model from overfitting to supervision with class prototypes, we introduce a novel self-supervised multiview consistency loss. This objective explicitly forces the fused descriptor to be predictive of descriptors from \textit{unseen} viewpoints of the same instance. This compensates for the potential biases of class-target losses and encourages the learning of intrinsic, view-invariant 3D semantic properties.
In summary, our main \textbf{contributions} are:
\begin{itemize}
\item A novel transformer-based multiview fusion model that  effectively combines information from diverse viewpoints.
\item A new self-supervised multiview consistency loss that improves convergence and generalization by enforcing consistency with unseen views.
\item {A flexible implementation compatible with various single-view descriptors that outperforms previous fusion baselines.}
\end{itemize}

%% file: figures/11_teaser.tex
\begin{figure*}[tp]
    \centering
    \includegraphics[width=\linewidth]{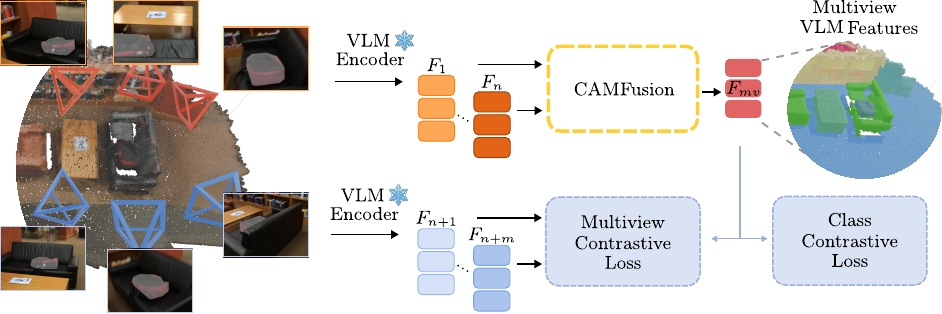}
    \caption{\textbf{\ours{} overview.} We propose a method to fuse vision-language descriptors from multiple views. Given masks of an object in $n$ images (red cameras), we extract per-view vision-language features ($F_1 \dots F_n$) and aggregate them using our CAMFusion to produce a unified descriptor $F_{mv}$. We also introduce a multiview contrastive loss that enforces consistency between the fused descriptor $F_{mv}$ and those from unseen viewpoints (blue cameras, $F_{n+1} \dots F_{n+m}$), complementing the supervised class contrastive loss typically used for this task.}
    \label{fig:teaser}
\end{figure*}

%% file: sections/02_realted_work.tex
\section{Related Work}
\noindent \textbf{VLMs} mainly rely on contrastive language-image pretraining, first introduced by~\cite{radford2021learning}, which aligns image and text representations within a shared latent space.
Building on this idea, subsequent works have refined both model architectures and training objectives. SigLIP~\cite{Zhai_2023_ICCV} replaced the computationally intensive softmax loss with a more efficient sigmoid-based contrastive objective.
EVA-CLIP~\cite{sun2023eva} further improved model performance through refined optimization strategies and training recipes.
SigLIP2~\cite{tschannen2025siglip} extended SigLIP by unifying several previously independent optimization techniques.
More recently, the Perception Encoder (PE)~\cite{bolya2025perception} introduced additional improvements in its optimization, achieving even stronger representations.
Although these models effectively map images and textual descriptions into a joint embedding space, they typically produce a single global descriptor per image.
Except for the PE, their visual encoders yield feature maps with limited spatial accuracy within the input image.

\subsubsection{Single-view open-vocabulary semantic} methods have leveraged VLMs to disentangle and segment the different semantic elements within a single image without requiring a predefined closed-vocabulary. 
These approaches can be broadly grouped into fine-tuned and training-free methods.
\textbf{Fine-tuned methods} achieve a strong performance by task-specific optimization with segmentation losses. MAFT+~\cite{jiao2024collaborative} fine-tunes a CLIP image encoder alongside a Mask2Former-based architecture~\cite{cheng2021mask2former}, using a transformer to condition text embeddings on image features. CAT-SEG~\cite{cho2024cat} jointly fine-tunes both image and text encoders, aggregating cost volumes across spatial locations and semantic classes using Swin Transformers~\cite{liu2021swin}. TAP~\cite{pan2024tokenize} follows a different direction, adapting a SAM-like architecture to predict semantic tokens by distilling knowledge from EVA-CLIP into a decoder, effectively unifying semantic segmentation and recognition.
\textbf{Training-free methods}, in contrast, leverage large pre-trained foundation models without further optimization. CaR~\cite{sun2024clip} adopts a recurrent approach that iteratively refines mask proposals by altering input images with visual prompts and filtering them using CLIP's~\cite{radford2021learning} similarity scores. 
Other approaches focus on enhancing feature maps: ProxyCLIP~\cite{lan2024proxyclip} uses a proxy attention module to fuse CLIP features with DINOv2~\cite{oquab2023dinov2} correlation maps, while CorrCLIP~\cite{zhang2024corrclip} leverages SAM2~\cite{ravi2024sam} masks to refine DINOv2 correlations and guide CLIP feature pooling.
Trident~\cite{shi2024harnessing} adopts a patch-based approach, combining CLIP and DINO patch features modulated by SAM’s final activations to obtain dense, per-patch descriptors, though at a significant overload.
Similarly, TextRegion~\cite{xiao2025textregion} integrates language-image descriptors from multi-scale patches, while using 2D segmentation masks for masked pooling and to compute per-mask vision-language descriptors.
While these training-free approaches yield strong visual–language representations, they remain limited to single-view scenarios and do not explicitly handle instances observed from multiple viewpoints.

\subsubsection{3D open-vocabulary semantic segmentation}
methods mostly rely on 2D segmentation masks to condition vision-language single-view descriptors, and then proceed to fuse them into a unified one.
OpenMask3D~\cite{takmaz2023openmask3d} computes single-view descriptors as the average of descriptors computed from multi-level masks, and then the multiview one as the average of all viewpoints. Open3DIS~\cite{nguyen2024open3dis} and Any3DIS~\cite{Nguyen_2025_CVPR} follow a similar approach, although iterating over how 2D masks are obtained.
OpenScene~\cite{peng2023openscene} also computes multi-level descriptors for each instance's viewpoint, fuses them with an average pooling, and uses them to distill a supervised instance segmentation network to also predict vision-language descriptors from 3D points.
HOV-SG~\cite{werby2024hierarchical} computes single-view descriptors by merging descriptors of a segmentation mask, its corresponding bounding box, and the whole image, and then selects the most representative descriptor based on their similarity to the average.
Similarly, OVO~\cite{ovo} also merges these descriptors, although using a small neural-network to predict per-dimensional weights. 
Expanding on it, OV3R~\cite{gong2025ov3r} introduces Dinov2~\cite{oquab2023dinov2} features into the neural-network to predict better weights.
Finally, OpenYolo3D~\cite{boudjoghra2025openyolo} first uses YOLO-World to, given the input set of classes, estimate 2D bounding boxes for each viewpoint. These predictions are matched to 3D instances predicted by Mask3D~\cite{hou2023mask3d}, and the final class is selected as the most common prediction from all viewpoints.
Note how these approaches rely on naive fusion mechanisms (\eg average or L1-Medoid), highlighting the need for a robust fusion to bridge the gap between rich 2D semantics and consistent 3D representations.

%% file: sections/03_method.tex
\section{Method}
\input{figures/31_method}
From a collection of 2D vision-language descriptors \(\mathcal{F} = \{F_1, ..., F_{n}\}\) extracted from 
\(N\) different viewpoints of the same 3D instance, our method predicts a fused multiview vision-language descriptor \(F_{mv}\).
We cast this fusion as a super-resolution problem, and solve it using a multiview transformer. In addition to the supervised contrastive objective used in prior work~\cite{ovo, gong2025ov3r}, we introduce a self-supervised multiview contrastive loss. We find that this self-supervised objective is essential for learning generalizable descriptors and for outperforming existing naive fusion strategies.
In this section, we first outline the standard framework for estimating 3D vision-language descriptors from multiple RGB-D views and review simple baselines for multiview aggregation. We then present our transformer-based fusion module, followed by a detailed description of the supervised and self-supervised training objectives.
\subsection{Multiview fusion of vision-language descriptors}
The standard pipeline for assigning open-vocabulary vision-language descriptors to a 3D instance reconstructed from a sequence $\mathcal{I} = \{\left(I_1, D_1\right), ..., \left(I_n, D_n\right)\}$ of $n$ posed RGB-D images begins in 2D.
Given this input, the first step is to use a foundation model \(\mathcal{E}_{mask}\) to generate image segmentation masks $\mathcal{M} = \{m_1, ..., m_n\}$ for a particular target instance (\eg the backpack on the sofa in our \cref{fig:teaser}) in every viewpoint where it is visible.
These masks can be obtained either by leveraging a 3D model to predict 3D instance masks from the backprojected pointcloud, or by using a 2D model (\eg SAM~\cite{Kirillov_2023_ICCV,ravi2024sam}) to generate image masks and establishing correspondences across viewpoints,
\begin{equation}
    \mathcal{M} = \mathcal{E}_{mask}\left(\mathcal{I}\right).
\end{equation}
Then, for each viewpoint, the images $\mathcal{I}$ are combined with the masks $\mathcal{M}$ to isolate instances, and extract their corresponding single-view vision-language descriptors
\begin{equation}
    \mathcal{F} 
    = \mathcal{E}_{img}\left(\mathcal{I}, \mathcal{M}\right),
\end{equation}
using the encoder \(\mathcal{E}_{img}\) of a vision-language model (\eg the CLIP image encoder).
This process is typically applied either to all $n$ viewpoints or only to the best viewpoints per 3D instance, yielding a set of noisy and view-dependent descriptors
$\mathcal{F} = \{F_1, ..., F_{n}\}$. 
As noted, despite research efforts on obtaining localized feature descriptors~\cite{zhang2024corrclip, xiao2025textregion,shi2024harnessing, cho2024cat}, these still struggle to perfectly isolate instance-specific semantics. Vision-language descriptors may be partial, ambiguous, or contaminated by background or clutter, which motivates the need for robust multiview fusion. The unified descriptor for the 3D instance is obtained by fusing the descriptors from all views as
\begin{equation}
    F_\text{mv} = \phi\left(\mathcal{F}\right),
\end{equation}
where \(\phi\) is typically implemented as a mean, average pooling, or medoid (either L1 or average) in the existing literature.
\subsection{Multi-view fusion as super resolution}
In this work, we reinterpret fusion not merely as an averaging operation, but as a semantic super-resolution problem. This formulation arises from two key observations about 3D open-vocabulary perception. 
First, the target instance is constant across viewpoints. While background, lighting, and occlusion may vary, the underlying semantic identity remains the same.
Second, different viewpoints provide complementary, rather than redundant, information. A rear view may capture a ``tail'', while a frontal view reveals a ``snout'', but neither view alone offers a complete semantic description (\eg ``dog''). Naively averaging these descriptors often dilutes specific features.
Instead, we aim to ``super-resolve'' a complete semantic descriptor by allowing each view to query complementary information from others.

Given the set of single-view descriptors \(\mathcal{F}\) of a 3D instance, in this work we draw inspiration from multiview geometry models~\cite{cabon2025must3r, liu2025slam3r} and employ a multiview transformer architecture \(\mathcal{T}\) to compute the fused descriptor
\begin{equation}
    F_{mv} = \mathcal{T}\left(\mathcal{F}\right).
\end{equation}
Our architecture, illustrated in \cref{fig:main_figure}, consists of an initial linear projection,
followed by \(d\) multiview attention blocks \(\{B_1,...,B_d\}\), and a final learned latent pooling stage.
At each multiview block \(B_d\), each feature vector performs self-attention and cross-attention to a memory \(M_n^d\) containing features of all other viewpoints from the previous block, followed by a linear layer, all with residual connections
\begin{equation}    
    E_i^d = B_d\left(E_i^{d-1}, M_i^{d}\right).
\end{equation}
At each block \(d\), the memory \(M_i^d\) of viewpoint \(i\) is made of the features generated by all \textit{other viewpoints} at previous block \(d-1\), \(M_i^d = \{E_1^{d-1}, ..., E_{n-1}^{d-1}\}\).
Finally, the fused multiview descriptor \(F_{mv}\) is computed with a latent pooling, a cross-attention operation where the final per-view features \(\{E_1^d,...,E_n^d\}\) act as keys and values, and a learned latent vector \(L\) serves as query
\begin{equation}
    F_{mv} = \mathrm{LatentPooling}\left(\{E_1^d,...,E_n^d\}, L\right).
\end{equation}
A single latent vector \(L\) is optimized during training and shared across instances.
\input{figures/32_loss}
\subsection{Self-Supervised Multi-View Contrastive Loss}
\label{sec:loss}
Current vision-language models (\eg CLIP, SigLIP) rely on a supervised contrastive loss. For each image-text pair, the loss forces their corresponding image and text embeddings to be similar to each other and different to the descriptors of all other samples in the batch.
Similarly to~\cite{ovo}, we use the sigmoid contrastive loss to train our models, forcing the fused descriptors to be similar to the text embedding of their corresponding ground-truth semantic class,
\begin{equation}
   \mathcal{L}_{c} =
   \frac{-1}{|\mathcal{B}|} \!\sum_{i=1}^{|\mathcal{B}|} \!\sum_{j=1}^{|\mathcal{B}|} \log \!\left(\! \frac{1}{1 \!+\! \exp(z_{ij}(-t \mathbf{F}_{mv}^i \!\cdot\! \mathbf{y}_j \!+\! b))}\! \right),
\end{equation} 
where $\mathcal{B}$ is the batch, $\mathbf{F}_{mv}^i$ is the descriptor predicted for instance $i$, $\mathbf{y}_j$ is the target text embedding of instance $j$, and $z_{ij}$ is $1$ for positive instance pairs, \ie \(i = j\), and $-1$ for negative ones, $i \ne j$.
While effective, this purely supervised approach can cause the model to overfit to the prototypical semantic classes, ignoring visual nuances crucial for robust 3D understanding and generalization.
To address this issue, we adapt the supervised contrastive loss to enable multiview self-supervision as
\begin{equation}
   \mathcal{L}_{mv} = 
   \frac{-1}{|\mathcal{B}||\mathcal{U}|} \!\sum_{i=1}^{|\mathcal{B}|} \!\sum_{j=1}^{|\mathcal{U}|} \log\!\left( \frac{1}{1 \!+\! \exp(z_{ij}(-t \mathbf{F}_\text{mv}^i \!\cdot\! \mathbf{F}_j \!+\! b))} \!\right)
\end{equation} 
where
\begin{equation}
    |\mathcal{U}| = \{F_{n+1}^1, ..., F_{n+m}^1\} \cup ... \cup \{F_{n+1}^b, ..., F_{n+m}^b\},
\end{equation}
is a set of $m$ single-view descriptors from unseen viewpoints of the same $b$ instances in $\mathcal{B}$.
The core idea is training the fused descriptor $\mathbf{F}_{mv}^i$ to be predictive of unseen viewpoints of the same instance $\mathbf{F}_j$, rather than just its class name.
The final loss is the direct summation of both components $\mathcal{L} = \mathcal{L}_{c} + \mathcal{L}_{mv}$.
%
To further improve generalization, we incorporate a semantic class mask into the loss (see \cref{fig:loss}). This mask modifies $z_{ij}$ such that it is $1$ if two instances belong to the same semantic class, \ie $\text{class}_i = \text{class}_j$.
This ensures that instances in the batch sharing the same semantic label are not treated as negative examples, preventing the model from incorrectly pushing away semantically similar descriptors.
We ablate in \cref{sec:ablations} the impact of our proposed loss and its components.

%% file: figures/31_method.tex
\begin{figure*}
    \centering
    \includegraphics[width=\linewidth]{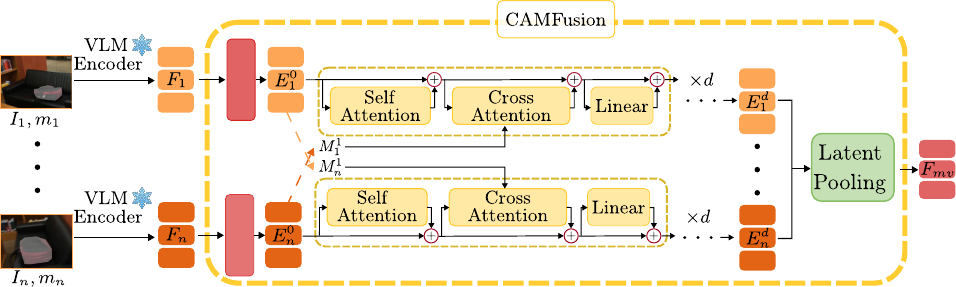}
    \caption{\textbf{\ours{} architecture.} Given a set of vision-language descriptors $F_1, ..., F_n$ of a 3D instance at $n$ views, these are processed by a multi-view transformer. At each block $d$, each embedding $E_i^d$ alternates between attending to it self and attending to its memory $M_i^d$ made of embeddings from other views. Finally, a learned latent pooling computes the final multi-view vision-language feature $F_{mv}$.}
    \label{fig:main_figure}
\end{figure*}

%% file: figures/32_loss.tex
\begin{figure}[b]
\centering
\begin{subfigure}[t]{0.49\linewidth}    
    \includegraphics[width=\linewidth]{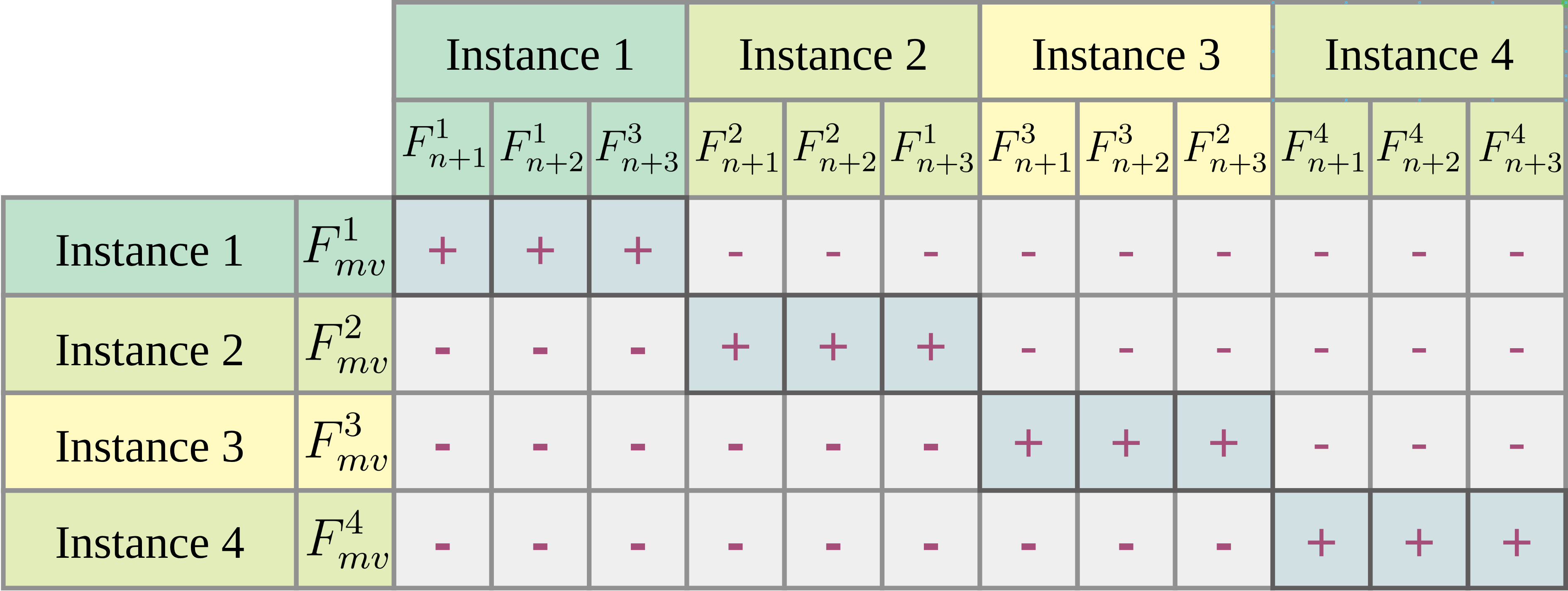}
    \caption{Multi-view contrastive  w/o class mask.}
\end{subfigure}
\begin{subfigure}[t]{0.49\linewidth}    
    \includegraphics[width=\linewidth]{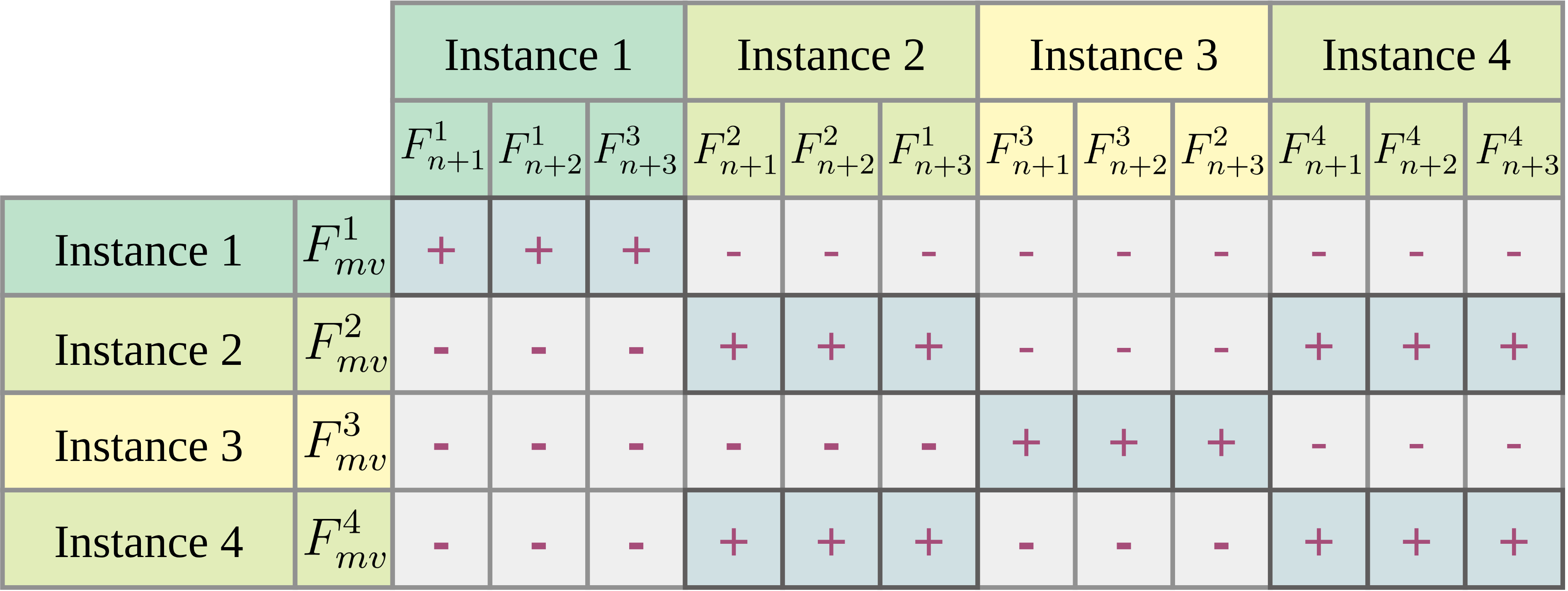}
    \caption{Multi-view contrastive  w/ class mask.}
\end{subfigure}
    \caption{\textbf{Multiview contrastive loss w/o (a) and w/ (b) the class mask.} The mask sets as positive target instances of the same class, preventing the contrastive loss from pushing their descriptors apart.}
    \label{fig:loss}
\end{figure}

%% file: sections/04_experiments.tex
\section{Experiments}
\label{sec:exp}
To assess generalization, we train on ScanNet++~\cite{yeshwanth2023scannetpp} and report test metrics on unseen datasets, specifically Replica~\cite{straub2019replica}, ScanNet~\cite{dai2017scannet}, and 3RScan~\cite{wald2019rio}.
First, using ground-truth 3D instance masks, we ablate our training choices in Replica, and then evaluate the performance of \ours\ with these masks. 
Finally, we integrate it with state-of-the-art methods for 3D semantic segmentation and instance segmentation.

\subsubsection{Implementation details.}
We rely on the PE ViT-L/14~\cite{bolya2025perception} with TextRegion~\cite{xiao2025textregion} to generate single-view descriptors.
\ours's inital projection layer consist of a linear layer with bias and dimension 1024 for both input and output.
The multiview transformer has 8 blocks of self-attention, cross-attention, and a linear block. Each attention layer has 8 heads with latent dimension 1024, pre-layer normalization, and GeLU activation functions. Each linear block is made of two layers with hidden dimension 4096, GeLU activation and bias.
When working with 3D masks as source (\ie ground-truth masks and Mask3D), we project the 3D masks to the image planes to generate 2D segmentation masks.
As we did not tackle 3D instance segmentation, we evaluate \ours\ using 2D/3D ground-truth instance masks, and 3D masks generated by two different pipelines, comparing the impact of our multiview fusion against their original approaches. 
We use:
\begin{itemize}
    \item OVO-mapping (OVO)~\cite{ovo}. An online open-vocabulary 3D segmentation pipeline. It uses SAM2 to segment input images, and tracks SAM2 masks over the sequence. The descriptor of each 3D instance is selected as the L1-medoid of the SigLip descriptors of the instance in every frame.
    \item Mask3D~\cite{hou2023mask3d}: A popular class-agnostic 3D instance segmentation network whose masks were also used by previous methods, like OpenYolo3D~\cite{boudjoghra2025openyolo}. 
\end{itemize}

\subsubsection{Training setup.} 
To computes single-view descriptors, we augment RGB images by applying random cropping, hue and saturation balance changes. Then to speed up training, the single-view descriptors of each 3D instance's viewpoint are precomputed once.
\ours\ is trained on a GPU Tesla V100, for 100 epochs, with AdamW, batch size 512, learning rate with a cyclic cosine decay schedule starting from \(0.1\times10^{-3}\) to \(0.5\times10^{-6}\) period of 7200 steps, and maximum learning rate decay of \(0.5\) on each cycle.
To compensate class unbalance, for each batch we sample with replacement 3D instances weighting by the inverse frequency of their semantic class on the dataset. Then, for each instance we randomly sample 10 viewpoints (5 as input and 5 as unseen targets) producing different observation of each instance on each epoch.
During evaluation, we select as input the 5 viewpoints with the best visibility following the same criteria as previous methods~\cite{ovo, peng2023openscene, nguyen2024open3dis}.

\subsubsection{Baselines.}
We evaluate \ours\ on 3D instances obtained from ground-truth, 2D (OVO), and 3D (Mask3D) input data, and compare it against semantic segmentation (OpenScene, HOV-SG, OV3R, OVO, and Open3DIS) and semantic instance segmentation (OV-3DIS, Open3DIS, Any3DIS) methods that produce CLIP-based representations as well as against an open-class classification baseline (Open-Yolo 3D).
For a fair comparison, we also report results for average pooling, as the most common multiview fusion strategy, using the same instance masks and single-view features (PE+TR+AvgPool).
When evaluating with OVO, we additionally include results using their L1-Medoid with our single-view features (OVO w/ PE+TR).
Finally, to ensure a fair comparison with the most recent OV-3DIS, we evaluate their fusion mechanism both with their original Alpha-CLIP features and with our single-view features (OV-3DIS w/PE+TR), using ground-truth 3D masks in both cases.

\subsubsection{Datasets.}
We train on ScanNet++ dataset, and evaluate it's generalization abilities on Replica, ScanNetv2, and 3RScan validation sets.
\textbf{ScanNet}\textbf{++}\textbf{v2} contains over 1K real indoor scenes, with ground-truth 3D meshes, high quality \(1752\times1168\) RGB images for training and validation, camera trajectory, and instance and semantic annotations for 3.4K classes. 
We use the official training set with 856 scenes and 2878 annotated classes for optimization.
\textbf{ScanNetv2} also captures real indoor scenes at RGB resolution of \(1296\!\times\!968\) and depth resolution of \(640\!\times\!480\).
It also has ground-truth 3D meshes with ground-truth instance and semantic annotations.
ScanNetv2 has two sets of annotations, the original set with 20 classes (ScanNet20), and an expanded set with 200 classes (ScanNet200)~\cite{rozenberszki2022language}.
We use the official validation set of 312 scenes. 
Image blur makes ScanNet more challenging than ScanNet++, and despite their common name, their data distributions are significantly different.
\textbf{Replica} is a synthetic dataset generated from high-fidelity real-world data.
Scenes consist of ground-truth 3D meshes with semantic annotations for 51 different classes. For all scenes, RGB-D sequences have been rendered at \(1200\!\times\!680\). For Replica we use the common 8 scenes subset (\textit{office-0...4}, \textit{room-0...2}) with NICE-SLAM camera trajectories~\cite{zhu2022nice}.
\textbf{3RScan} is a real-world dataset with image size of \(960\times540\), low-resolution depth images and rendered depth maps, and providing ground-truth 3D instance semantic segmentation for 528 classes.
We use their validation split of 157 scenes.

\subsubsection{Metrics}
Despite 3D semantic and semantic instance segmentation tackling similar problems, they are usually evaluated with different metrics~\cite{jung2025details, peng2023openscene, ovo}.
For 3D semantic segmentation, we measure mean Intersection over Union and Accuracy (IoU and Acc). Following previous methods~\cite{engelmann2024opennerf,werby2024hierarchical,ovo} we report them weighted by frequency (f-IoU and f-Acc) on Replica, and split by \textit{head}, \textit{common}, and \textit{tail} tertiles in ScanNet20 and ScanNet200.
For 3D semantic instance classification we report mean Average Precision (mAP) at masks overlap threshold of \(25\%\), \(50\%\), and the average across the range \(\left[50\%:95\%\right]\) with increments of \(5\)\%.
Following OV-3DIS~\cite{jung2025details}, we report results for both single-prediction per instance (Top-1), and top-600 predictions of higher confidence (Top-k).
Note that semantic segmentation metrics account for background classes (``wall'', ``floor'', and ``ceiling''), while instance metrics do not (except on 3RScan).
Following previous methods, we also report instance classification metrics split by \textit{head} ($AP_h$), \textit{common} ($AP_c$), and \textit{tail} ($AP_t$), tertiles based on classes frequency on each dataset.
Furthermore, in 3RScan we also report the performance on its subset of classes not present in ScanNet++ class set. Despite 173 of the 528 classes being new, only 30 of them are actually observed in the validation set.
Finally, we highlight in each table the \colorbox{Green!25}{\textbf{first}}, \colorbox{SpringGreen!45}{second}, and \colorbox{Yellow!30}{third} best results
%
%
\input{figures/41_qualitative_replica.tex}
\input{figures/43_ablation_views}
\subsection{Ablations} \label{sec:ablations}
We report an ablation in Replica, using ground-truth 3D instances, of the different components of our system trained on ScanNet++, see \cref{tab:ablation}.
We start with the same contrastive class loss introduced by \cite{ovo}, although relying on a weighted resampling strategy to compensate for class imbalance, instead of their proposed per-class weights.
While the weighted resampling plays a key role attenuating class imbalance, our new  multiview self-supervised contrastive loss is the key to outperform previous naive approaches, specially when using novel views  as target, \textit{contrastive loss} and \textit{mv contrastive loss} in \cref{tab:ablation}.
Adding the class-mask to the losses improves 3D instance classification in exchange to a slow drop in performance in 3D semantic segmentation metrics.
Note that although if naive approaches achieve a higher $mAP$ on tail categories, the small number of its instances makes it a really noisy metric.
{Finally, we evaluated the number of input views in \ours\ comparing the model trained with 5 views against the base average pooling in both Replica and ScanNet200. \ours\ performs better independently of the number of views, and benefits from multi view information.}
\input{tables/ablation}
\input{tables/sem_replica.tex}
\input{tables/sem_scannet200.tex}
\subsection{3D semantic segmentation}
When integrating \ours\ with OVO for online 3D semantic segmentation (\cref{tab:replica_sem}), our multiview fusion strategy increases OVO's performance from \(27\%\) to \(38\%\). Notably, only half of this gain comes from a better single-view descriptor (OVO w/ PE+TR at \(33\%\)).
As a result, \ours\ surpasses the current state-of-the-art, OV3R, by \(8\) percentage points (pp), including improvement on \textit{tail} classes where OVO with \textit{PE+TR} performs worse.
Note that OV3R relies on the same 3D instance predictions as OVO, differing only in its stronger single-view model.
In ScanNet (\cref{tab:scannet_sem}), \ours\ again outperforms OVO's fusion when using the same features on both ScanNet20 and ScanNet200 label sets. Also observe that OpenScene achieves substantially better performance on ScanNet20, although it does not extrapolate to ScanNet200, where its metrics degrade sharply {without generalizing to less common categories}.
\input{tables/gt_replica_scannet}
\input{tables/gt_3rscan}
\input{tables/ins_replica}
\input{tables/ins_scannet200}
\subsection{3D Instance classification}
Evaluating \textbf{with ground-truth 3D masks}, the gap between \ours\ and previous methods is substantial (\cref{tab:replica_ins_gt} and \cref{tab:r3scans_ins_gt}).
On both ScanNet200 and Replica, we surpass OV-3DIS and AvgPool by more than \(10\) pp on Top-k and by over \(+5\) pp on Top-1. 
Moreover, \ours\ consistently achieves better metrics on \textit{tail} categories.
On 3RScan we also outperform previous baselines on most categories. For the subset of labels unseen during our training, \(\text{mAP}_\text{new}\), our model underperforms on Top-1 but achieves the best result in the Top-k setting, indicating that the generated descriptors generalize well to novel text labels despite the noisy nature of this small subset.
Regarding previous baselines, OV-3DIS and OpenYolo3D fall significantly behind both our \ours\ and the simpler AvgPooling.
This can also be observed in the qualitative results in \cref{fig:qualitative}. 
\ours\ consistently outperforms previous baselines disentangling better instances' semantics from their environment.
Overall, the large performance margin observed with ground-truth masks demonstrates that our method offers substantial headroom for improving future class-agnostic 3D instance segmentation pipelines. 
Finally, \cref{tab:replica_ins} and \cref{tab:scannet_ins} report results for \ours\ using the 3D instance masks predicted by Mask3D~\cite{hou2023mask3d}, the instance segmentation model used by OpenYolo3D. 
Although the performance gap is smaller than with ground-truth, \ours\ still clearly outperforms both average pooling (using the same single-view descriptors) and baselines like Open-YOLO 3D, Open3DIS, Any3DIS, and Search3D on both Top-1 and Top-k metrics.
Importantly, Open3DIS, Any3DIS, and Open-Yolo3D rely on a 2D ``grounded open-vocabulary'' detector that requires prior knowledge of the scene's classes at optimization, which limits their open-vocabulary capability.
Compared with the more recent OV-3DIS, our multiview fusion does outperform their Alpha-CLIP based descriptors for Top-1 predictions for both Replica and ScanNet200 and is only outperformed for Top-k on ScanNet200, despite being better on tail categories.
As OV-3DIS, on ScanNet200 we even outperform the closed-vocabulary version of Mask3D by \(1.5\) pp.
Finally, although OV-3DIS outperforms \ours\ on ScanNet200 for the Top-k setup, this is largely attributable to its higher-quality predicted 3D instance masks. This is confirmed by our ground-truth–mask experiments, where \ours\ performs significantly better when evaluated with the same masks.
Therefore, the stronger masks predicted by OV-3DIS would likely further improve performance when integrated with \ours.
Our results in the appendix highlight the importance of using strong single-view descriptors.
As a consequence, the fact that our model can be trained on a single GPU in under one day makes it easy to adapt to increasingly powerful vision–language encoders.
\subsection{Discussion}
While \ours\ demonstrates strong fusion performance, it also presents some limitations. First, as a pure descriptor fusion module, \ours\ is fully dependent on upstream 2D or 3D instance segmentation masks. Consequently, any segmentation errors propagate directly through the pipeline. Second, our current strategy for selecting input viewpoints relies on simple heuristics (e.g., mask size) or random sampling during training, which does not ensure that the most informative or complementary views are chosen. A more involved viewpoint-selection mechanism could substantially improve both efficiency and accuracy. Finally, our class-contrastive supervision uses only standard class names, with limited language augmentation (e.g., synonyms, LLM-generated descriptions, affordances). This may constrain the model’s ability to handle more complex open-vocabulary queries.

%% file: figures/41_qualitative_replica.tex
\begin{figure}[t]
    \centering

    \includegraphics[width=\linewidth]{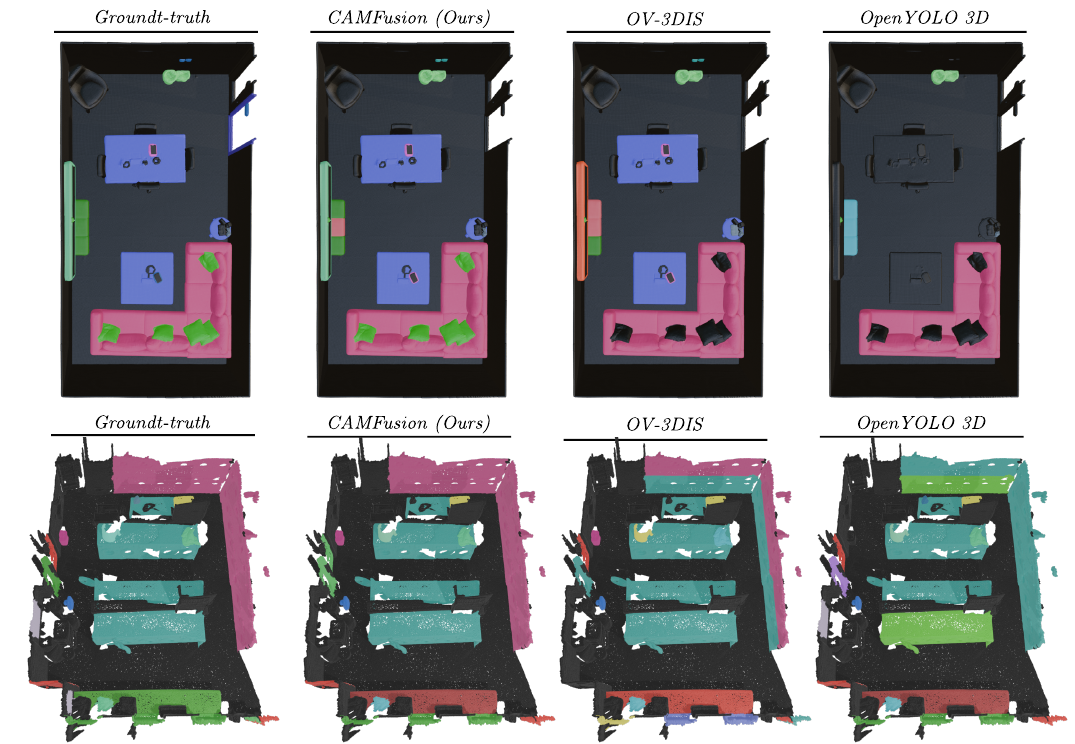}  

    \caption{Qualitative results for open-vocabulary semantic instance classification on \textbf{Replica} (top) and \textbf{ScanNet200} (bottom), using GT instance masks for all methods. We visualize results of our \ours\ against the ground-truth and the baselines OV-3DIS and Open YOLO 3D. Observe how \ours\ produces sharper and more coherent object boundaries filtering out the segmentation noise observed in the baselines, resulting in 3D masks that closely match the ground truth.}
    \label{fig:qualitative}
\end{figure}

%% file: figures/43_ablation_views.tex
\begin{figure}[h!]
\centering
\includegraphics[width=\linewidth]{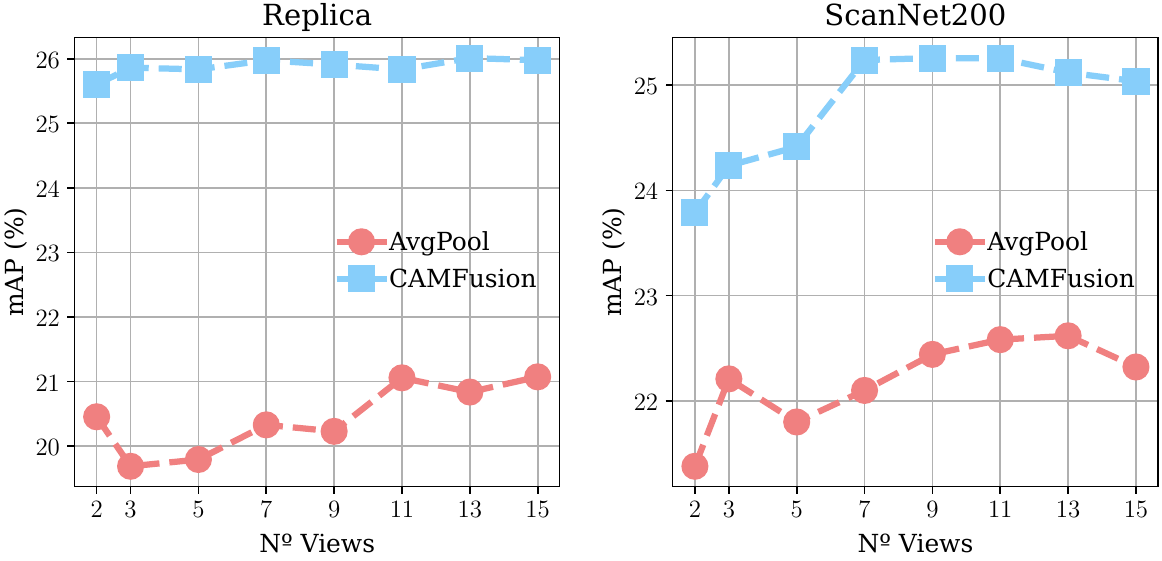}

\caption{\textbf{3D Instance segmentation vs number of views}. CAMFusion trained with 5 views and evaluated in Replica and ScanNet200 using Mask3D's masks.}
\label{fig:views}
\vspace{-0.1cm}
\end{figure}

%% file: tables/ablation.tex
\begin{table}[t]
\caption{\textbf{Ablation of loss components on Replica.} We compare naïve fusion strategies (Average Pooling, L1 Medoid) against our \ours, incrementally validating the impact of our weighted resampling and multiview constrastive loss.}
\label{tab:ablation}
\centering
\setlength\tabcolsep{2.6pt}
\tablesize
\begin{tabular}{lcccccccc} 
\toprule 
 MV method &   IoU &  Acc && mAP & mAP$_h$ & mAP$_c$ & mAP$_t$ \\ 
\midrule 
\small\textit{\textbf{Naive fusion baselines}} \\
AvgPool & 55.4 & \rd 69.5 & & 64.2 & 67.6 & 61.8 & \st63.3\\
L1 Medoid &52.2 & 68.6 & & 64.1 & 66.9 & 62.6 & \nd62.7 \\
\midrule  
\small\textit{\textbf{Our fusion method}} \\
class loss  &  31.4 & 43.7  && 37.3 & 55.3 & 34.8 & 21.8\\ 
+ weigh. resampling &  51.9 & 63.7 && 58.7 & 65.6 & 57.8 & 52.6 \\ 
~+ contrastive loss & \rd 57.1 & 65.8 && 64.4 & 76.0 & \nd 64.4 & 53.0\\ 
~~+ mv contrastive loss & \st61.9 &\st 73.1 && \nd67.2 & \nd 78.2 & \rd 64.0 & 59.4 \\ 
~~~+ class mask \textbf{(final)} & \nd 58.7 & \nd 70.7 && \st68.9 & \st81.0 & \st 66.0 & \rd 59.8 \\ 
\bottomrule
\end{tabular}
\end{table}

%% file: tables/sem_replica.tex
\begin{table}[h!]
\caption{\textbf{Open-Vocabulary 3D Semantic Segmentation on Replica.} \ours\ integrated with OVO outperforms all previous approaches by a significant margin, even against the same single-view descriptors (``OVO w/ PE+TR'').}
\label{tab:replica_sem}
\centering
\setlength{\tabcolsep}{2.pt}
\tablesize
\begin{tabular}{lcc|cccccc}
   \toprule
   & \multicolumn{2}{c}{All} & \multicolumn{2}{c}{Head} & \multicolumn{2}{c}{Common} & \multicolumn{2}{c}{Tail} \\
  \cmidrule(lr){2-3} \cmidrule(lr){4-5} \cmidrule(lr){6-7} \cmidrule(lr){8-9}
  Method &  IoU & Acc & IoU & Acc & IoU & Acc & IoU & Acc \\ \midrule
  OpenScene(OSeg)~\cite{peng2023openscene} & 15.9 & 31.7 & 35.4 & 45.6 & 20.1 & 31.3 & 5.8 & 17.6 \\ 
  HOV-SG~\cite{Werby-RSS-24}                           &  22.5 & 34.2 & 35.9 & 44.2 &  23.6 &  42.3 & 8.0 & 16.1  \\
  Open3DIS~\cite{nguyen2024open3dis} (SigLip) & 25.6 & 38.7 &  49.7 &  64.4 & 22.1 &  42.4 & 4.9 & 9.4 \\ 
  OV3R~\cite{gong2025ov3r} & \rd 30.7 & 41.3&  43.1 & 55.3 & \rd 27.1 & 38.0 &\nd 21.8 & \rd30.6\\
  OVO~\cite{ovo}  &  27.0 &  39.1 & \rd 45.0 & \rd 59.9 &  25.1 & \rd 38.5 &  11.0 & 18.8   \\
  OVO~\cite{ovo} w/ PE+TR & \nd 33.0 & \nd 53.4 & \nd 53.3 & \st 67.5 & \nd27.6 & \nd 48.8 & \rd 18.0 & \nd44.1 \\
  \textbf{OVO w/ ours} & \st 38.3 & \st56.5 &\st53.9 & \nd66.3 & \st38.0 & \st56.7 & \st22.9 & \st46.5 \\ 
  \bottomrule
\end{tabular}
\end{table}

%% file: tables/sem_scannet200.tex
\begin{table}[t]
\caption{\textbf{Open-Vocabulary 3D Semantic Segmentation on ScanNet 20 and 200}. OVO with \ours\ significantly improves on the challenging ScanNet200 vocabulary.}
\label{tab:scannet_sem}
\centering
\setlength{\tabcolsep}{1.3pt}
\tablesize
\begin{tabular}{lccccccccc}
   \toprule
   & \multicolumn{4}{c}{ScanNet20} && \multicolumn{4}{c}{ScanNet200}   \\
  \cmidrule(lr){2-5}  \cmidrule(lr){6-10}  
  Method &  IoU & Acc & f-IoU & f-Acc &&  IoU & Acc & f-IoU & f-Acc \\ \midrule
  OpenScene(LSeg)~\cite{peng2023openscene} & \st 52.7 & \nd 65.5 & \st 67.7 & \st 79.6 && 8.1 & 12.0 & \st 52.0 & \st 66.9 \\
  OpenScene(OSeg)~\cite{peng2023openscene} & \nd47.0 & \st70.3 & \nd57.7 & \nd 69.8 && \rd11.6 & 22.8 & 24.5 & 29.2 \\
  Open3DIS~\cite{nguyen2024open3dis}& 24.7 & 40.9 & 32.5 & 45.3  &&9.4 & 17.0 & 22.9 & \rd32.2 \\
  OVO w/ PE+TR~\cite{ovo} & 31.8 & 61.2 & 49.1 & 63.3 && \nd 16.0 & \nd42.8 & 38.5 &  49.9 \\
  \textbf{OVO w/ ours} & \rd 35.7 & \rd 62.5 & \rd 54.5 & \rd 68.3 && \st17.7 & \st45.3 & \nd45.2 & \nd56.8\\ 
  \bottomrule
\end{tabular}
\end{table}

%% file: tables/gt_replica_scannet.tex
\begin{table}[h!]
\caption{\textbf{3D Instance Segmentation on Replica and ScanNet200 using ground-truth 3D masks}. } 
\label{tab:replica_ins_gt}
\centering
\setlength{\tabcolsep}{.5pt}
\tablesize
\begin{tabular}{clccccccccc}
   \toprule
   && \multicolumn{4}{c}{Replica} && \multicolumn{4}{c}{ScanNet200}  \\ \cmidrule{3-6} \cmidrule{8-11}
  &Method &  mAP & mAP$_h$ & mAP$_c$ & mAP$_t$ && mAP & mAP$_h$ & mAP$_c$ & mAP$_t$ \\
  \midrule
 \multirow{5}{*}{\rotatebox[origin=c]{90}{Top-k}}
 & OpenYolo3D ~\cite{boudjoghra2025openyolo} & 44.0 & 47.5 & 49.0 & 35.5 &&39.6&  43.3&   36.8&   38.5\\ 
 & OV-3DIS~\cite{jung2025details} & 44.3 & 46.9 & 47.2 & 38.7 &&30.9 &33.9 & 26.5 & 32.5 \\
 & OV-3DIS w/ PE+TR & \rd 52.3 & \rd 67.3  & \rd 47.8  & \rd 45.4 &&  \rd46.6 & \rd51.1 & \rd38.9 & \rd50.5   \\
 & AvgPool  w/ PE+TR & \nd 63.1 & \nd 74.2 & \nd 62.4 & \nd 52.8 && \nd52.3 & \nd57.7 & \nd42.3 & \nd58.0 \\ 
 & \textbf{Ours} & \st73.7 & \st84.2 & \st81.0 & \st55.9  && \st62.1 & \st70.3 &\st 51.2 &\st 65.6  \\
 \midrule
 \multirow{5}{*}{\rotatebox[origin=c]{90}{Top-1}}
  &OpenYolo3D ~\cite{boudjoghra2025openyolo}  & 46.2 & 51.6 & 48.8 & 38.1 && 39.6 & 43.3 & 36.9 & 38.5 \\ 
 & OV-3DIS~\cite{jung2025details} & 39.1 & 42.9 & 39.3 & 35.2 && 30.3 & 30.3 & 25.8 & 36.0  \\
 & OV-3DIS w/ PE+TR & \rd 55.6& \rd 57.8 & \rd 49.8 & \rd 59.5 && \rd 45.1 & \rd43.7 & \rd38.7 & \rd54.5 \\
  & AvgPool w/ PE+TR & \nd 64.2 & \nd 67.6 & \nd61.8 & \st63.3 &&  50.0\nd &\nd 51.9 & \nd41.4 &\nd 57.9\\
  &\textbf{Ours} & \st68.9 & \st81.0 & \st66.0 & \rd 59.8 && \st 56.0 & \st60.9 & \st46.7 & \st61.3\\ 
  \bottomrule
\end{tabular}
\end{table}

%% file: tables/gt_3rscan.tex
\begin{table}[h!]
\caption{\textbf{3D Instance Segmentation on 3RScan using ground-truth 3D masks}. \ours\ achieves the overall best performance despite the blurry input images.}
\label{tab:r3scans_ins_gt}
\centering
\setlength{\tabcolsep}{3.5pt}
\tablesize
\begin{tabular}{clccccccccc}
   \toprule
  &Method & mAP & mAP$_h$ & mAP$_c$ & mAP$_t$ & mAP$_{new}$  \\
  \midrule
 \multirow{5}{*}{\rotatebox[origin=c]{90}{Top-k}}
 & OpenYolo3D ~\cite{boudjoghra2025openyolo} & 26.9 & 18.9 & \rd 25.7 & 28.9 &  17.1 \\ 
 & OV-3DIS~\cite{jung2025details} & 16.7 &  \rd 33.4 & 12.1 & 17.1 & 8.1 \\
 & OV-3DIS w/ PE+TR & \rd 30.1 &  32.2 & \rd25.7 & \rd32.7 & \rd19.7  \\
 & AvgPool  w/ PE+TR & \nd 37.1 & \nd 34.6 & \nd36.7 & \nd37.8 & \nd30.3  \\ 
 & \textbf{Ours} & \st 42.9 & \st 35.0 & \st39.3 & \st46.5 & \st31.8  \\
 \midrule
 \multirow{5}{*}{\rotatebox[origin=c]{90}{Top-1}}
  &OpenYolo3D ~\cite{boudjoghra2025openyolo}  & 17.6 & 13.3 & 17.7 & 18.2 & 12.3 \\ 
 & OV-3DIS ~\cite{jung2025details}& 16.0 & 33.5 & 12.7 & 15.5 & 8.9  \\
 & OV-3DIS w/ PE+TR & \rd 23.6 & \nd 33.8 & \nd 21.6 & \rd 23.2 & \nd 25.6  \\
  & AvgPool w/ PE+TR  & \nd 23.8 & \st 36.5 & \nd 22.3 & \rd 22.8 & \st 27.2\\
  &\textbf{Ours} & \st 30.5 & \nd 23.7 & \st 27.6 & \st 33.5 & \rd 14.5 \\ 
  \bottomrule
\end{tabular}
\end{table}

%% file: tables/ins_replica.tex
\begin{table}[ht!]
\caption{\textbf{3D Instance Segmentation on Replica.} We integrate \ours~with Mask3D.}
\label{tab:replica_ins}
\centering
\setlength{\tabcolsep}{0.1pt}
\tablesize
\begin{tabular}{clccccccc}
   \toprule
  &\multirow{2}{*}{Method} &3D mask&\multirow{2}{*}{ mAP }& \multirow{2}{*}{mAP$_{50}$} & \multirow{2}{*}{mAP$_{25}$} & \multirow{2}{*}{mAP$_h$ }& \multirow{2}{*}{mAP$_c$ }& \multirow{2}{*}{mAP$_t$ }\\
  & & proposals  & & & & & & \\
  \midrule
 \multirow{6}{*}{\rotatebox[origin=c]{90}{Top-k}}
  &\cellcolor{shadecolor}Open3DIS~\cite{nguyen2024open3dis} & \cellcolor{shadecolor} 2D+3D &\cellcolor{shadecolor}  18.5 &\cellcolor{shadecolor}  24.5 & \cellcolor{shadecolor} 28.2&\cellcolor{shadecolor}  - &\cellcolor{shadecolor} - &\cellcolor{shadecolor} -  \\
  &\cellcolor{shadecolor}OV-3DIS~\cite{jung2025details} & \cellcolor{shadecolor}  2D+3D & \cellcolor{shadecolor}  25.7 & \cellcolor{shadecolor}  34.9 &  \cellcolor{shadecolor} 42.3 & \cellcolor{shadecolor} -&\cellcolor{shadecolor} -&\cellcolor{shadecolor} - \\ \cdashline{2-9}
  & OV-3DIS~\cite{jung2025details} & 3D & \rd 18.9 & \rd 24.4 & \rd 32.1  & -&-&- \\
  & Open-Yolo3D~\cite{boudjoghra2025openyolo} &3D& 17.7 & 22.6 & 29.3 & \rd 18.5 & \rd 23.1 & \rd 11.6\\
  & \textbf{PE+TR+AvgPool} & 3D & \nd20.8 & \nd26.7 &\nd 39.0 &\nd 23.6 &\nd 22.2 &\nd 16.5\\
  & \textbf{Ours}  & 3D & \st 26.4 & \st 34.7 & \st 46.6 & \st 25.6 & \st 36.2 & \st 17.5\\ 
  \midrule
 \multirow{5}{*}{\rotatebox[origin=c]{90}{Top-1}}
 &\cellcolor{shadecolor}OV-3DIS~\cite{jung2025details} &\cellcolor{shadecolor} 2D+3D & \cellcolor{shadecolor} 22.6 &\cellcolor{shadecolor}  31.7 &\cellcolor{shadecolor}  37.7 & \cellcolor{shadecolor} - & \cellcolor{shadecolor} - & \cellcolor{shadecolor} -  \\\cdashline{2-9}
  & OV-3DIS~\cite{jung2025details} & 3D & \rd22.0 & \rd 26.7 & 32.5  & -&-&- \\
  &Open-YOLO3D~\cite{boudjoghra2025openyolo} & 3D & \nd 23.7 & \nd 28.6 & \nd 34.8 & \rd 19.0 & \st 35.9 & \nd 16.0 \\ 
  &\textbf{PE+TR+AvgPool} &3D &19.8 & 24.8 & \rd 33.2 & \nd 21.2 & \rd 25.0 &\rd 13.2 \\
  &\textbf{Ours} & 3D& \st 25.8 & \st 33.7 & \st 40.0 & \st 25.3 & \nd 35.4 & \st 16.9  \\
  \bottomrule
\end{tabular}
\end{table}

%% file: tables/ins_scannet200.tex
\begin{table}[h!]
\caption{\textbf{3D Instance Segmentation on ScanNet200.} Thanks to its novel instance segmentation, OV-3DIS is the only approach that competes with \ours.}
\label{tab:scannet_ins}
\centering
\setlength{\tabcolsep}{.4pt}
\tablesize
\begin{tabular}{clccccccc}
   \toprule
  &\multirow{2}{*}{Method} & 3D mask &\multirow{2}{*}{mAP } & \multirow{2}{*}{mAP$_{50}$} & \multirow{2}{*}{mAP$_{25}$} & \multirow{2}{*}{mAP$_h$ }& \multirow{2}{*}{mAP$_c$ }& \multirow{2}{*}{mAP$_t$ }\\
  & & proposal & & & & & & \\
  \midrule
  &\cellcolor{shadecolor}Mask3D~\cite{hou2023mask3d} (c.v.) & \cellcolor{shadecolor} 3D & \cellcolor{shadecolor}26.9& \cellcolor{shadecolor} 39.8 & \cellcolor{shadecolor} 21.7 & \cellcolor{shadecolor} 17.9 & \cellcolor{shadecolor} - & \cellcolor{shadecolor} - \\ \midrule
 \multirow{7}{*}{\rotatebox[origin=c]{90}{Top-k}}
 &\cellcolor{shadecolor}Open3DIS~\cite{nguyen2024open3dis} &\cellcolor{shadecolor}  2D+3D &\cellcolor{shadecolor}23.7&\cellcolor{shadecolor}  29.4 &\cellcolor{shadecolor}  32.8 & \cellcolor{shadecolor} 27.8 & \cellcolor{shadecolor} 21.2 &\cellcolor{shadecolor}  21.8 \\
 & \cellcolor{shadecolor}Any3DIS~\cite{Nguyen_2025_CVPR} &\cellcolor{shadecolor}  2D+3D &\cellcolor{shadecolor}25.8&\cellcolor{shadecolor}  - &\cellcolor{shadecolor}  - &\cellcolor{shadecolor}   27.4 &\cellcolor{shadecolor}   23.8 & \cellcolor{shadecolor} 26.4 \\
 & \cellcolor{shadecolor}OV-3DIS~\cite{jung2025details} &\cellcolor{shadecolor}  2D+3D &\cellcolor{shadecolor}32.7 &\cellcolor{shadecolor}  41.4 &\cellcolor{shadecolor}  45.3 & \cellcolor{shadecolor} 34.5 & \cellcolor{shadecolor} 30.7&\cellcolor{shadecolor}  33.1 \\ \cdashline{2-9}
 & OV-3DIS~\cite{jung2025details} & 3D & \st 29.0 & \nd37.6 & \nd 42.8 & \st 33.0 & \st 28.1 & \cellcolor{shadecolor} \nd25.3 \\
 & Open-YOLO3D~\cite{boudjoghra2025openyolo} & 3D & \rd 24.7 & \rd 31.7 &  36.2 & \rd 27.8 & \nd 24.3 & 21.6 \\ 
 & \textbf{PE+TR+AvgPool} & 3D & 21.8 & 31.0 & \rd 37.5 & 23.4 & 18.4 & \rd24.1 \\
 & \textbf{Ours} & 3D &\nd 27.6 &  \st 39.3 &  \st46.2 &  \nd31.8 & \rd 22.2 &  \st29.1 \\
 \midrule
 \multirow{6}{*}{\rotatebox[origin=c]{90}{Top-1}} 
 &\cellcolor{shadecolor}OV-3DIS~\cite{jung2025details} & \cellcolor{shadecolor} 2D+3D & \cellcolor{shadecolor}25.8 &\cellcolor{shadecolor}32.5 &\cellcolor{shadecolor}36.2 & \cellcolor{shadecolor}26.3 & \cellcolor{shadecolor}23.2 & \cellcolor{shadecolor}28.2  \\ \cdashline{2-9}
 & OV-3DIS~\cite{jung2025details} & 3D & \nd 24.2 & \nd31.8 & \rd36.4 & \nd 27.2 & \st 22.3 & \rd 23.1 \\
 & Search3D~\cite{takmaz2025search3d} & 3D & \rd 23.0 & - & - & \rd26.3 & \nd 21.2 & 21.4 \\ 
 &Open-Yolo3D~\cite{boudjoghra2025openyolo} &3D & 21.9 & 28.3& 31.7& 25.6& \rd 21.0& 18.5 \\
 & \textbf{PE+TR+AvgPool}&3D & 21.8 & \rd 31.0 & \nd 37.5 & 23.4 & 18.4 & \nd 24.1 \\
 & \textbf{Ours} &3D & \st 24.4 & \st 35.0 & \st 41.0 & \st 27.4 & 20.4 & \st 25.6 \\
  \bottomrule
\end{tabular}
\end{table}

%% file: sections/10_conclusion.tex
\section{Conclusion} \label{sec:conclusion}
In this paper, we propose \ours, a multiview transformer that generates multiview vision-language descriptors by fusing single-view descriptors of different viewpoints.
To optimize this model, we introduce a new self-supervised multiview consistency loss, that forces the multiview descriptors to be close to the single-view descriptors from novel unseen viewpoints.
Our experiments show that \ours~outperforms previous fusion methods on both 3D semantic segmentation and 3D instance classification tasks, both using oracle and estimated 3D proposals.
Furthermore, we believe that our multi-view loss could also be leveraged to supervise future feed-forward 3D semantic models, without being explicitly limited to multiview fusion.

%% file: sections/11_appendix.tex
\appendix
\setcounter{figure}{0} 
\setcounter{table}{0}
\setcounter{section}{0}
\renewcommand{\thefigure}{A-\arabic{figure}} 
\renewcommand{\thetable}{A-\arabic{table}}   
\section{Additional Ablations}
\label{sec:appendix}
\subsection{Single-View Descriptor Analysis}
To isolate the impact of the input quality on the final 3D representation and select the best descriptor for our model, we first evaluated various strategies for extracting single-view Language-Image descriptors.

\noindent \textbf{Baselines.} We compared five extraction strategies across two backbones (SigLIP-SO400M/16 and Perception Encoder ViT-L/14):
\begin{itemize}
\item \textbf{Standard Cropping:} We evaluate \textit{Mask Crop} (masking the background and centering the object) and \textit{BBox Crop} (cropping to the instance bounding box, retaining context).
\item \textbf{HOV-SG \cite{werby2024hierarchical}:} A heuristic ensemble method that alpha-blends descriptors from the whole image, the bounding box, and the masked crop using predefined weights and softmax normalization.
\item \textbf{Weights Predictor:\cite{ovo}} A learned variant of HOV-SG that uses a lightweight neural network to predict the fusion weights for the three image views (Global, Box, Mask).
\item \textbf{TextRegion \cite{xiao2025textregion}:} Applies the segmentation mask directly to the VLM's final feature map to guide attention pooling, avoiding image resizing artifacts.
\end{itemize}
 We utilized ground-truth 3D instance masks projected into all camera viewpoints to generate the single-view descriptors.
 \begin{table}[h]
    \centering
    \small
    \setlength{\tabcolsep}{1.5pt} 
    \begin{tabular}{lcccc|ccccc|cccccc}
        \toprule
                        & \multicolumn{4}{c}{Replica} & &\multicolumn{4}{c}{ScanNet200} & & \multicolumn{4}{c}{ScanNet++}\\\cline{2-5} \cline{7-10} \cline{12-16}
                        & IoU & Acc & f-IoU & f-Acc && IoU & Acc & f-IoU & f-Acc&& IoU & Acc & f-IoU & f-Acc \\ \hline
        \textit{\small\textbf{SigLip SO400M/16}} & & & & && & & & && & & &  \\
    MaskCrop          & 25.8 & 46.3 & 31.1 & 50.8 && 14.4 & 26.8 & 19.4 & 36.1 && 19.7 & 39.6& \nd 48.4& \rd 57.0   \\
    BBoxCrop          & 22.1 & 33.7 & 36.9 & 55.1 &&14.6 & 22.9 & 23.6 & 35.9 && 18.5 & 49.0 & 29.7& 43.2    \\
    HOV-SG            & 30.1 & \rd 53.1 &  46.1  & 63.8  &&  18.5 & 32.4 & 27.1 & 42.5 && 26.1 & 53.3 & 41.0 & 53.9\\
    TextRegion &  33.6 &  45.7 &  46.0   & \rd 63.7   && \nd 22.2 & \nd 34.9 & 25.1 & 37.6 && \nd 29.1 & \st 58.4 &  42.7 & 54.4 \\
    Weights Predictor       & \rd 33.7 & \nd 57.2 & 66.6 \nd & 77.5 \nd&& \rd 21.8 & \st 39.1 & \nd 45.1 & \nd 60.8 && \st 41.7 & \rd 53.6 & \st 72.3 & \st 81.6   \\
    \midrule
    \textit{\small\textbf{PE-Core L }} & & & & && & & & && & & &  \\
    MaskCrop          & 25.8 & 35.2 & 42.6   & 59.3 && 14.9 & 23.5 & 23.3 & 36.4  && 19.9 & 51.7& 38.0 & 49.3   \\
    BBoxCrop          & 14.3 & 23.6 & 26.8   & 36.0 && 9.4 & 16.3 & 16.2 & 23.5  && 10.3 & 34.9 & 14.9 & 25.1   \\
    HOV-SG            & 22.4 & 36.4 & 37.0  & 51.9 && 15.5 & 24.2 & 23.6 & 34.0 && 20.2 & 54.0 & 26.8 & 39.5   \\
    TextRegion       & \st 42.1 & \rd 51.8 & \st 67.1  & \st 79.1.9 && \st 24.8 & \rd 34.1 & \rd 40.0 & \rd 54.1 && \rd 26.9 & \st 58.4 & \rd 45.3 & \nd 57.3  \\
    \bottomrule
    \end{tabular}
    \caption{Single-view 2D semantic segmentation metrics on Replica, ScanNet200, ScanNet++ using ground-truth segmentation masks.}
    \label{tab:sv_2d}
\end{table} 
\noindent \textbf{Results.} \cref{tab:sv_2d} reports 2D classification metrics on Replica, ScanNet200, and ScanNet++. TextRegion and the Weights Predictor consistently outperform standard cropping methods. Notably, TextRegion exhibits a significant performance boost when paired with the Perception Encoder (PE) compared to SigLIP. We attribute this to the PE's feature map retaining more localized semantic information, which aligns well with TextRegion's mask-pooling mechanism.

\subsection{Cross-Evaluation of Naive Fusion Strategies}
We investigated the cross-impact between single-view feature quality and naive multiview fusion algorithms. 
We utilized single-view descriptor from previous ablation, which are fused using three naive baselines: Average Pooling, L1 Medoid, and Cosine Similarity Medoid.

\noindent \textbf{Results.}
As shown in \cref{tab:mv_sv_replica}, when relying on naive fusion, the final 3D performance is dominated by the quality of the single-view input rather than the specific fusion strategy. The best-performing single-view methods (Weights Predictor and TextRegion) consistently yield the best 3D results regardless of the fusion method used.

Interestingly, simple Average Pooling consistently outperforms the Medoid selection approaches when coupled with high-quality inputs like TextRegion + PE. As a consequence we selected it as the main naive-fusion baseline on the main experiments of the paper. 
This suggests that while naive fusion can provide a performance boost over single-view estimation (reducing noise and improving metrics) it remains fundamentally limited by the input quality and do not leverage multi-view information.
These results highlight the importance of our CAMFusion approach, which actively learns to synthesize improved features rather than passively aggregating them.
\begin{table}[h]
    \centering
    \setlength\tabcolsep{1.0pt}
    \begin{tabular}{lcccccccccccccc} 
    \toprule 
      & \multicolumn{4}{c}{Replica} &&\multicolumn{4}{c}{ScanNet200} &&\multicolumn{4}{c}{ScanNet++}\\\cline{2-5} \cline{7-10} \cline{12-15}
    Method & IoU & Acc & f-IoU & f-Acc && IoU & Acc & f-IoU & f-Acc&& IoU & Acc & f-IoU & f-Acc \\ 
    \midrule 
    \midrule 
    \multicolumn{6}{l}{\textit{\small\textbf{SigLIP-SO400M}}} \\ 
    \midrule
    \multicolumn{6}{l}{\textit{\small\textbf{MV Fusion: AvgPooling}}} \\ 
      BboxCrop & 30.9 & 46.8 &  60.9 & 71.0 && 25.1 & 38.6 & 32.3 & 43.2 && 24.9 & 41.9 & 36.9 & 51.2 \\ 
      MaskCrop & 30.3 & 43.2 & 62.8 & 66.2 && 12.7 & 18.4 & 17.8 & 22.6 &&  24.0 & 32.7 & 52.3 & 59.6 \\ 
      HOVSG & 32.9 & 43.9 & 65.7 & 73.8  && 24.3 & 33.9 & 32.5 & 41.9 && 32.6 & 48.4 & 48.6 & 62.2\\ 
      TextRegion & \st{45.5} & \st{65.2} &  68.9 & 76.9  && \st 34.0 & \st{52.9} & 37.6 & 46.4 && 36.7 & \st 60.3 & 53.2 & 65.3\\ 
      WeightsPredictor & 41.5 & 49.4 &  \st{75.4} & \st{85.5}  && 18.3 & 25.1 &  \rd{39.1} & \rd{50.1} && \nd38.5 & 45.1 & \st 68.3 & \st 80.4\\ 
    \midrule
    \multicolumn{6}{l}{\textit{\small\textbf{MV Fusion: L1Medoid}}} \\ 
      BboxCrop & 34.0 & 49.7 & 55.4 & 66.5&& 24.4 & 40.6 & 29.2 & 40.1  &&  21.1 & 39.8 & 35.4 & 49.4 \\ 
      MaskCrop & 31.4 & 47.9 & 60.7 & 65.7 &&  15.1 & 22.2 & 19.8 & 25.6 && 20.6 & 31.2 & 46.6 & 54.6 \\ 
      HOVSG & 34.4 & 46.4 & 62.8 & 70.2 && 24.2 & 36.3 & 31.4 & 41.4 && 29.1 & 44.6 & 44.0 & 58.2  \\ 
      TextRegion &  \nd{41.1} &  \rd{60.9} & 62.6 & 70.9 && \nd{32.3} & \st{52.9} & 37.7 & 46.6 && 33.2 & \rd{58.6} & 50.4 & 63.4\\  
      WeightsPredictor & 39.9 & 47.3 & \nd{70.8} & \nd{81.8} && 22.9 & 31.3 & \st{40.1} & \st{51.7} && \rd 38.4 & 46.2 & \rd 67.7 & \rd 79.4 \\ 
     \midrule
     \multicolumn{6}{l}{\textit{\small\textbf{MV Fusion: CosSimMedoid}}} \\ 
     BboxCrop & 31.3 & 48.2 & 53.9 & 65.5 && 23.2 & 39.1 & 28.7 & 39.4 &&  20.7 & 39.8 & 35.3 & 50.2 \\ 
      MaskCrop & 31.7 & 46.5 & 57.1 & 61.5 && 13.4 & 20.6 & 18.4 & 24.0 &&  19.7 & 29.8 & 46.7 & 54.5 \\ 
      HOVSG & 35.4 & 47.1 & 63.8 & 70.6 && 23.0 & 34.9 & 30.6 & 40.5 &&  29.5 & 44.6 & 43.9 & 58.1 \\ 
      TextRegion & \rd{40.4} & \nd{62.4} & 64.0 & 72.3 &&  \rd{31.9} &  \rd{52.1} & 36.7 & 45.7  && 34.2 & \nd 59.0 & 54.3 & 66.1 \\  
      WeightsPredictor & 38.0 & 46.2 & \nd{70.4} & \rd{81.6} && 21.8 & 30.0 &  \nd{39.5} & \nd{50.9}  && \st 39.0 & 47.6 & \nd 67.8 & \nd 79.5  \\ 
     \midrule
     \midrule
    \multicolumn{6}{l}{\textit{\small\textbf{PE Core L14 336}}} \\ 
    \midrule
    \multicolumn{6}{l}{\textit{\small\textbf{MV Fusion: AvgPooling}}} \\ 
      BboxCrop & 21.5 & 41.6 & 28.2 & 44.8 && 17.8 & 36.4 & 18.5 & 28.2 && 16.9 & 34.4 & 25.7 & 36.9   \\ 
      MaskCrop & 40.2 & 56.5 & 70.4 & 76.9 && 28.9 & 43.9 & 36.7 & 46.5 &&  33.0 & 52.7 & \st{53.5} & \st{63.0} \\ 
      HOVSG & 31.7 & 50.2 & 50.7 & 64.8 && 27.3 & 48.0 & 32.0 & 43.6 && 30.6 & 51.2 & 38.6 & 51.7 \\ 
      TextRegion & \st 58.9 & \st 72.0 & \st 81.4 & \st 88.6 && \nd 33.9 & \nd 55.7 & \st 44.0 & \st 55.2 && \st{35.4} & \st 59.4 & \nd50.9 & 62.9\\ 
     \midrule 
     \multicolumn{6}{l}{\textit{\small\textbf{MV Fusion: L1Medoid}}} \\ 
      BboxCrop & 21.5 & 43.6 & 26.8 & 42.6 && 15.3 & 34.0 & 16.0 & 25.1 &&  13.6 & 32.9 & 19.5 & 30.1\\ 
      MaskCrop & 34.5 & 54.8 & 59.3 & 70.0&&  21.1 & 39.0 & 29.0 & 39.2  &&  23.6 & 46.4 & 46.5 & 56.9 \\ 
      HOVSG & 28.3 & 49.9 & 45.6 & 60.1 && 22.9 & 42.5 & 27.7 & 39.0 &&  23.3 & 47.2 & 31.3 & 44.4 \\ 
      TextRegion & \rd 54.0 & \rd 67.0 & \nd 75.9 & \nd 85.1 & & \rd 33.3 & \st 56.0 & \st 44.0 & \st 55.2 && \rd{31.1} & \rd{55.9} & \rd{50.2} & \rd{62.2}  \\ 
     \midrule 
     \multicolumn{6}{l}{\textit{\small\textbf{MV Fusion: CosSimMedoid}}} \\ 
      BboxCrop & 17.8 & 40.1 & 26.6 & 43.1 && 14.3 & 32.5 & 15.5 & 24.5 && 13.3 & 31.7 & 19.3 & 29.5  \\ 
      MaskCrop & 36.4 & 58.8 & 56.2 & 68.0 && 20.6 & 38.4 & 28.3 & 38.6 &&  23.0 & 46.2 & 46.9 & 57.2  \\  
      HOVSG & 29.6 & 49.4 & 48.4 & 62.4 && 22.9 & 42.5 & 27.5 & 38.7 && 23.7 & 47.8 & 30.9 & 44.2  \\
      TextRegion & \nd 54.2 & \nd 67.5 & \rd 75.6 & \rd 84.6&&  \rd{32.6} & \rd 55.0 & \rd 43.7 & \rd 54.9 && \nd{31.2} & \nd{57.3} & \rd{50.2} & \nd{62.3}  \\ 
    \bottomrule
    \end{tabular}
    \caption{Cross multiview/single-view methods per class instance classification on Replica using GT 3D instances and PE-Core-L14-336 and SigLIP-So400M descriptors. Highlighted in \textbf{bold} the best from each multi view subgroup, and in \colorbox{colorFst}{first}, \colorbox{colorSnd}{second}, and \colorbox{colorTrd}{third} best overall.}
    \label{tab:mv_sv_replica}
    \end{table}
\section{Additional qualitative results}
\begin{figure*}[h]
    \centering
    \includegraphics[width=0.9\linewidth]{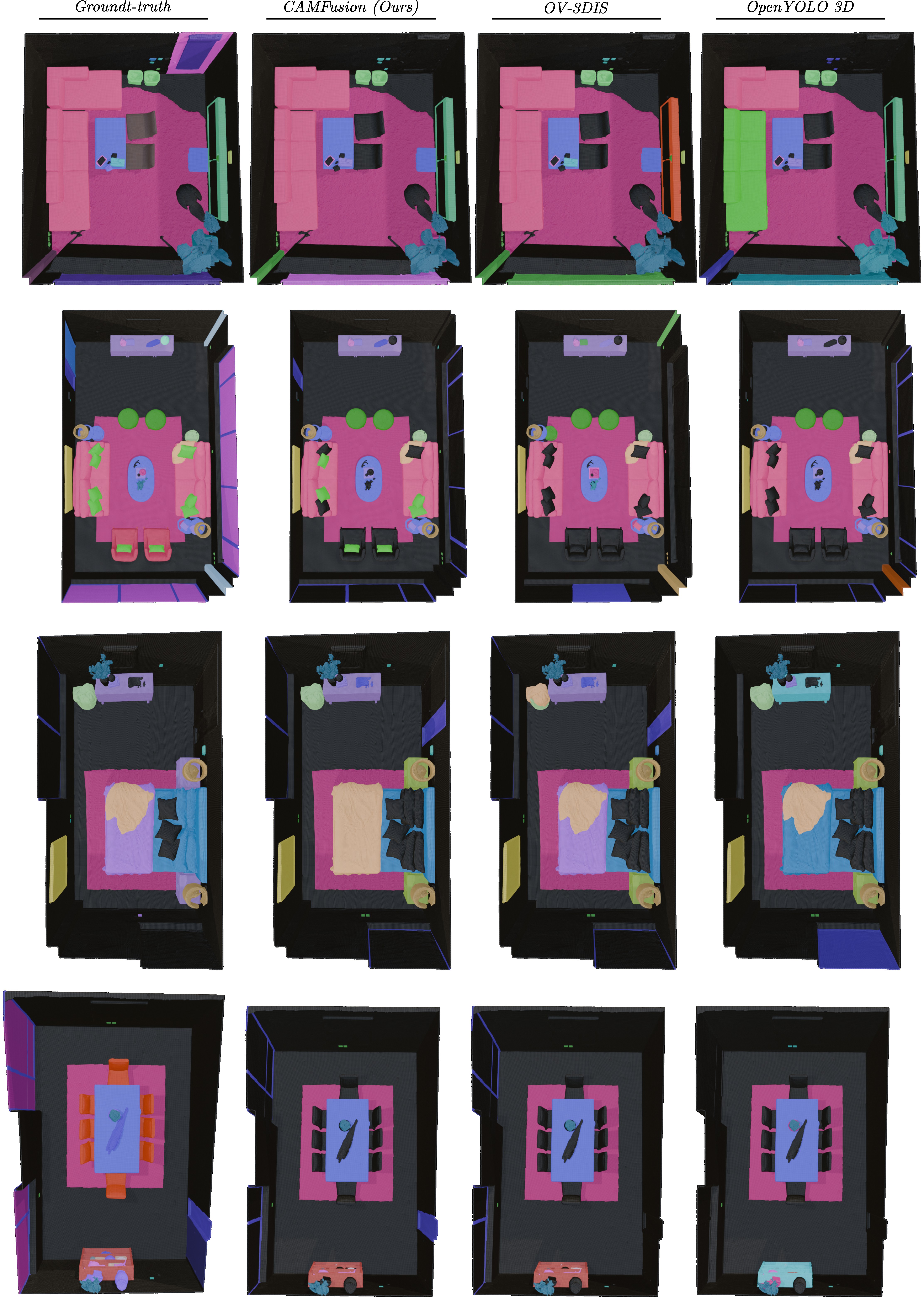}
    \caption{Qualitative comparison of 3D instance semantic classification with ground-truth 3D masks on Replica.}
    \label{fig:sup_replica}
\end{figure*}

\begin{figure*}[h]
    \centering
    \includegraphics[width=0.9\linewidth]{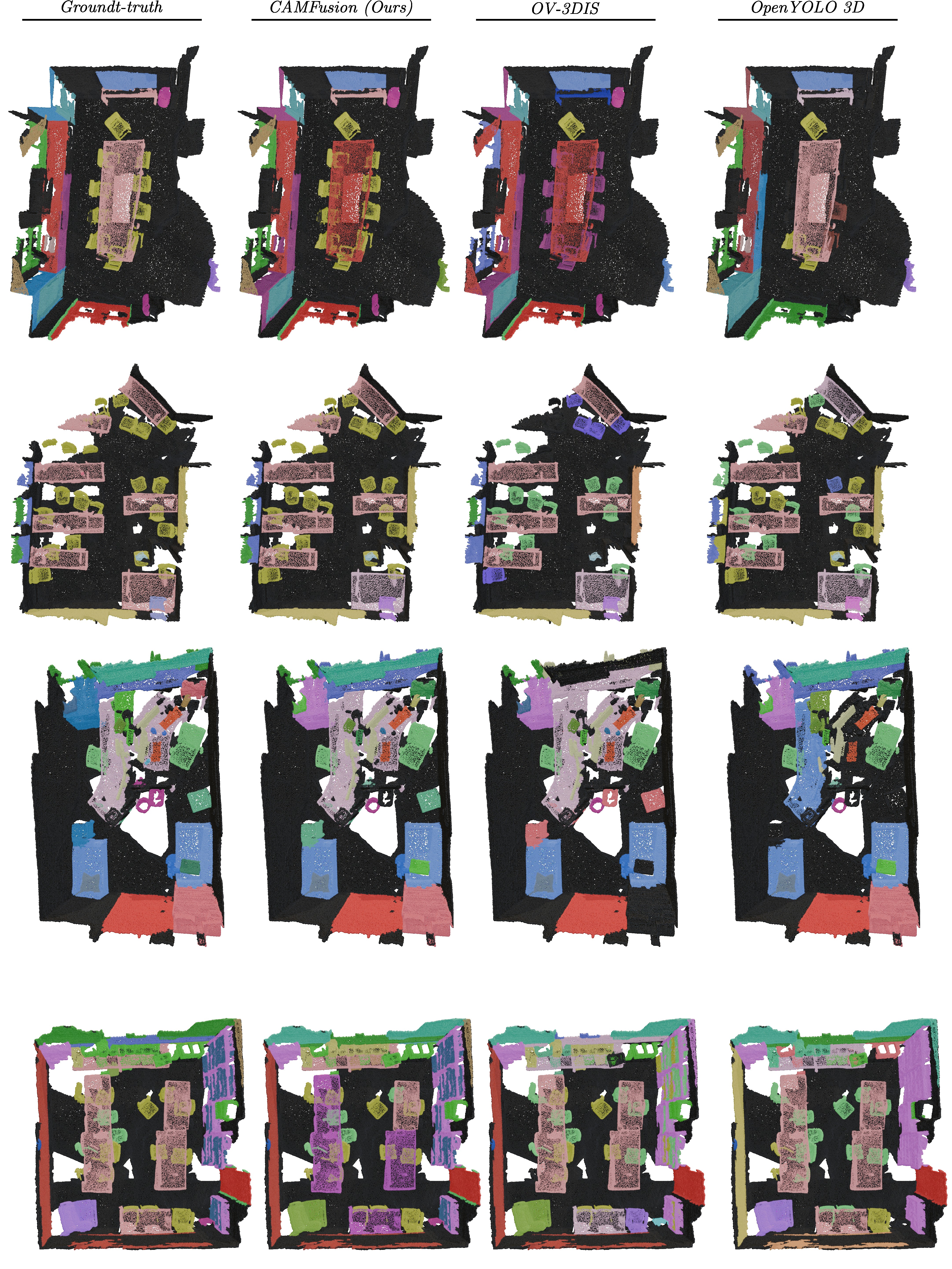}
    \caption{Qualitative comparison of 3D instance semantic classification with ground-truth 3D masks on ScanNet200.}
    \label{fig:sup_scannet}
\end{figure*}

We include in \cref{fig:sup_replica} and \cref{fig:sup_scannet} additional qualitative visualizations on instance classification from ground-truth instance masks in Replica and ScanNet200.
